\documentclass[conference, letterpaper]{IEEEtran}
\usepackage[top=0.72in, bottom=1in, left=0.62in, right=0.62in]{geometry}

\IEEEoverridecommandlockouts
\usepackage{amsmath}
\usepackage{graphicx}
\usepackage{cite}
\usepackage{amsmath,amssymb,amsfonts, amsthm}
\usepackage[linesnumbered,ruled]{algorithm2e}
\usepackage{algorithmicx}
\usepackage{algpseudocode}
\usepackage{graphicx}
\usepackage{textcomp}
\usepackage{xcolor}
\usepackage{xspace}
\usepackage{soul}
\usepackage{kotex}

\usepackage{tcolorbox}
\usepackage{float}  
\usepackage{caption} 

\usepackage{booktabs} 
\usepackage[normalem]{ulem} 
\usepackage{ragged2e}

\usepackage{xspace}
\newcommand{\BfPara}[1]{{\noindent\bf#1.}\xspace}
\usepackage{tabularx}
\usepackage{color}
\definecolor{pgreen}{rgb}{0,0.5,0}
\definecolor{pred}{rgb}{0.9,0,0}
\definecolor{ppurple}{rgb}{0.5,0,0.5}

\usepackage{subfigure}
\usepackage{xcolor}
\usepackage{tikz}
\usepackage{listings}
\usepackage{verbatim}

\usepackage[T1]{fontenc}
\usepackage[utf8]{inputenc}

\definecolor{color1}{RGB}{228,26,28}
\definecolor{color2}{RGB}{55,126,184}
\definecolor{color3}{RGB}{77,175,74}
\definecolor{color4}{RGB}{152,78,163}

\usepackage{float}

\definecolor{promptbg}{HTML}{F2F2FF}      
\definecolor{promptborder}{HTML}{808080}  

\usepackage{amsmath,amsfonts,amssymb}
\usepackage{tabularx}
\usepackage{tikz}

\begin{document}
\title{Hallucination-Aware Generative Pretrained Transformer for Cooperative Aerial Mobility Control}

\author{\IEEEauthorblockN{
    Hyojun Ahn$^{\dag}$, 
    Seungcheol Oh$^{\dag}$, 
    Gyu Seon Kim$^{\dag}$, 
    Soyi Jung$^{\ddag}$, 
    Soohyun Park$^{\S}$, and 
    Joongheon Kim$^{\dag}$
}
\IEEEauthorblockA{
$^{\dag}$Department of Electrical and Computer Engineering, Korea University, Seoul, Republic of Korea \\
$^{\ddag}$Department of Electrical and Computer Engineering, Ajou University, Suwon, Republic of Korea \\
$^{\S}$Division of Computer Science, Sookmyung Women's University, Seoul, Republic of Korea \\
E-mails: 
\texttt{hyojun@korea.ac.kr}, 
\texttt{seungoh@korea.ac.kr}, 
\texttt{kingdom0545@korea.ac.kr}, \\
\texttt{sjung@ajou.ac.kr}, 
\texttt{soohyun.park@sookmyung.ac.kr},
\texttt{joongheon@korea.ac.kr} 
}
}

\maketitle
\begin{abstract}
This paper proposes \textit{SafeGPT}, a two-tiered framework that integrates generative pretrained transformers (GPTs) with reinforcement learning (RL) for efficient and reliable unmanned aerial vehicle (UAV) last-mile deliveries. In the proposed design, a \textit{Global GPT} module assigns high-level tasks such as sector allocation, while an \textit{On-Device GPT} manages real-time local route planning. An RL-based safety filter monitors each GPT decision and overrides unsafe actions that could lead to battery depletion or duplicate visits, effectively mitigating hallucinations. Furthermore, a dual replay buffer mechanism helps both the GPT modules and the RL agent refine their strategies over time. Simulation results demonstrate that SafeGPT achieves higher delivery success rates compared to a GPT-only baseline, while substantially reducing battery consumption and travel distance. These findings validate the efficacy of combining GPT-based semantic reasoning with formal safety guarantees, contributing a viable solution for robust and energy-efficient UAV logistics.


\end{abstract}

\section{Introduction}\label{sec:intro}

Unmanned aerial vehicles (UAVs) have gained significant attention as a promising solution to last-mile logistics due to their strong capability for rapid and autonomous deliveries. To ensure practical deployment in such scenarios, effective route planning that accounts for UAV battery life preservation is crucial~\cite{10144378}. However, designing such battery-efficient UAV route planners is highly nontrivial, as the dynamic nature of UAV environments poses significant challenges for widely used optimization solutions (e.g. mixed-integer programming) which are usually designed under static assumptions \cite{Dorling2017}. To cope with the dynamic environment, reinforcement learning (RL) techniques have been applied to train UAVs to learn routing policies from interaction with the environment \cite{10543150}. While RL-based planners offer adaptability and can outperform static heuristics in complex scenarios, they require extensive training, tuning, and generally lack formal safety guarantees. 

This paper considers a problem setting where battery-efficient UAV route planning is solved using a two-tiered generative pretrained transformers (GPTs), which have emerged as a powerful paradigm for UAV planning. In this setting, global-GPT determines the sector assignments for subsets of UAVs, each equipped with an on-device GPT. Subsequently, on-device GPTs generate routing path for their corresponding UAVs while accounting for battery consumption constraints. Unlike conventional optimization methods, the GPT-based framework effectively handles the dynamic and uncertain nature of UAV environments by modeling complex sequential dependencies. Compared to reinforcement learning, GPTs require less environment-specific tuning and training, yet offer strong generalization and decision-making capabilities, which makes them a promising alternative for adaptive UAV route planning.


Although GPT offers significant advantages over conventional optimization methods and RL for UAV route planning, the use of GPT introduces several new challenges. First, there is the issue of hallucination, a well-known characteristic of GPT, which may lead to undesirable outcomes, such as UAVs making duplicate visits to customers. Second, since GPT is a pre-trained model, it is difficult to fine-tune to our specific problem structure. 
To address these challenges, we propose SafeGPT framework which integrates RL–based safety filters to override hallucinated or unsafe actions, and employs a dual replay buffer mechanism that allows two-tiered GPT modules to reference past experiences for continuous refinement of routing decisions. This hybrid approach unifies the generalization strengths of GPT with the adaptive, real-time learning capabilities of reinforcement learning, ensuring that UAV routes remain both energy-efficient and compliant with strict battery consumption and operational constraints in dynamic environments.

\begin{figure*}[t]
    \centering
    \includegraphics[width=\textwidth]
    {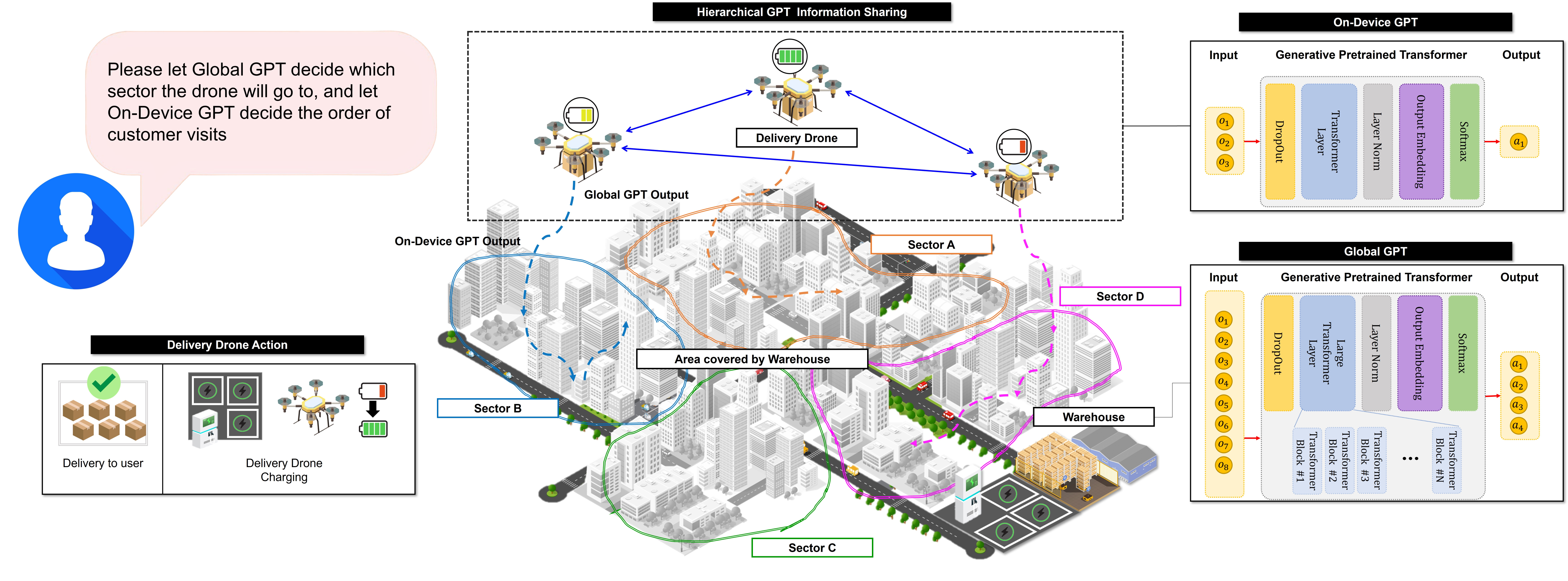}
    \caption{A conceptual overview of SafeGPT in a drone delivery scenario.}
    \label{fig:overview}
    \vspace{-5mm}
\end{figure*}

\BfPara{Related Work}\label{sec_2:Preliminaries}
LLMs have been introduced as promising candidates for autonomous planning in robotics. For instance, Ahn et al.  have demonstrated how GPT suggestions can be grounded in robotic affordances using pretrained skill value functions~\cite{brohan2023doascan}, while other studies have augmented LLM planners with symbolic planning modules to ensure logical consistency \cite{liu2021symbolic}. Despite these advances, LLM-based planners may still produce hallucinated or unsafe actions due to insufficient environmental grounding \cite{Chen2023}.
To mitigate these issues, recent research has integrated safety filters and RL techniques into the planning process. Khan et al. proposed SAFER, which pairs an LLM planner with a dedicated safety agent for real-time risk assessment~\cite{khan2021safer} , and Yang et al. introduced a "safety chip" that enforces constraints such as linear temporal logic~\cite{yang2023safetychip}. In addition, Xu et al. employed a learned critic to preempt hallucinations~\cite{xu2022saferl}, and constrained policy optimization via RL from human feedback as well as fine-tuning methods like Safe-RL have been shown to reduce high-risk behaviors~\cite{achiam2017constrained}.

\BfPara{Contributions}\label{sec:contributions}
The main contributions are as follows, i.e., 
\textit{\textbf{GPT-Driven Planning with RL}} (A novel framework integrating a GPT-based planner for context-aware scheduling decisions with an RL safety filter to detect and correct hallucinated outputs, synthesizing semantic reasoning with formal safety assurances), \textit{\textbf{Dual Replay Buffer}} (A mechanism that stabilizes the RL agent's learning process while providing memory for GPT during planning inference, creating a synergistic feedback loop that helps avoid past planning errors), and \textit{\textbf{Case Study in UAV Smart Logistics}} (Validation in a realistic UAV-based logistics scheduling scenario, demonstrating the RL component's effectiveness in mitigating unsafe actions and achieving robust delivery performance under dynamic conditions)

\section{Modeling} \label{sec:Modeling}
\subsection{UAV Networks and Energy Modeling}

A reference network for an unmanned package delivery system is established, as illustrated in Fig.~\ref{fig:overview}. The network comprises a collection of \(N\) drones, denoted as \(D_i\) for each \(i \in \mathcal{N}\), and a single warehouse, denoted as \(W\). The warehouse serves as the central hub covering the entire urban area, from which drones depart to proceed directly to their designated delivery zones for package distribution. Moreover, the warehouse is outfitted with wireless energy charging capabilities to restore the energy of drones upon landing. In light of the limited energy capacity of the drones, the system imposes a restriction that each drone carries no more than \(4\) packages concurrently.

\begin{table}[t!]
\centering
\scriptsize
\caption{Specification of UAV model\cite{joby_uam}.}
\renewcommand{\arraystretch}{1.0}
\begin{tabular}{l||r}
\toprule[1pt]
\textsf{\textbf{Notation}} & \textsf{\textbf{Value}} \\ \midrule
Maximum number of package, $\Lambda$ & 4 \\
Flight speed, $v$ & 73.762\,[$\mathrm{m/s}$] \\
Aircraft mass including battery and propellers, $m$ & 1,815\,[$\mathrm{kg}$]\\
Aircraft weight including battery and propellers, ${W}={mg}$ & 17,799\,[$\mathrm{N}$]\\ 
Rotor radius, $R$ & 1.45\,[$\mathrm{m}$] \\
Rotor disc area, $A=\pi R^{2}$ & 6.61\,[$\mathrm{m^{2}}$] \\
Number of blades , $b$ & 5 \\
Rotor solidity, $s=\frac{0.2231b}{\pi R}$ & 0.2449 \\
Blade angular velocity, $\Omega$ & 78\,[$\mathrm{radius/s}$]\\
Tip speed of the rotor blade , $U_{tip}=\Omega R^{2}$ & 112.776\,[$\mathrm{m/s}$] \\
Air density, $\rho$ & 1.225\,[$\mathrm{kg/m^{3}}$] \\
Fuselage drag ratio, $d_{0}=\frac{0.0151}{sA}$ & 0.01 \\
Mean rotor-induced velocity in hovering, $v_{0}=\sqrt\frac{W}{s\rho A}$ & 26.45\,[$\mathrm{m/s}$] \\
Profile drag coefficient, $C_{d}$ & 0.045 \\
Incremental correction factor to induced power, $k$ & 0.052 \\
\bottomrule[1pt]
\end{tabular}
\label{tab:parameters of uam}
\vspace{-5mm}
\end{table}

To accurately represent the energy constraints of UAVs, an energy consumption model based on aerodynamic power calculations is formulated with the parameters in Table~\ref{tab:parameters of uam}. During the hovering phase—critical for take-off and landing while transporting payloads—the power expenditure \(P_h\) is defined as,
    $P_h=\underbrace{\frac{C_{d}}{8}\rho sA\Omega^3R^3}_{\textit{blade profile, }P_{o}}+\underbrace{(1+k)\frac{W^{\frac{3}{2}}}{\sqrt{2\rho A}}}_{\textit{induced, }P_{i}}$. 
In addition, for forward propulsion during transit, the propulsion power consumption can be mathematically expressed as following formulation,
$P_p= \underbrace{P_i\left(\sqrt{1+\frac{v^4}{4v_0^4}}-\frac{v^2}{2v_0^2}\right)^{0.5}}_{\textit{induced}}+ \underbrace{P_0\left(1+\frac{3v^2}{U_{tip}^2}\right)}_{\textit{blade profile}}+\underbrace{\frac{1}{2}d_0\rho sAv^3}_{\textit{parasite}}$,
where \(v\) designates the cruising speed at the altitude, \(v_0\) is the mean rotor-induced velocity, and \(U_{tip}\) represents the tip speed of the rotor blade. 
Concurrently, the battery system is characterized by its own parameters. Each drone is assigned a battery capacity of \(150\,kWh\) and is configured to receive a maximum charge of \(30\,kWh\) per journey. The charging operation requires approximately \(5\) minutes per journey using a charger with a supply power of \(360\,kW\), yielding a per-journey charge rate of \(20\%\) and achieving a full state of charge (SOC) in roughly \(25\) minutes.
Integrating the aerodynamic energy consumption model with the battery specifications establishes a comprehensive framework for simulating UAV operations within an energy-constrained context. This modeling approach facilitates the evaluation of both operational efficiency and safety in an unmanned package delivery system.


\begin{figure*}[t]
    \centering
    \includegraphics[width=0.65\textwidth]{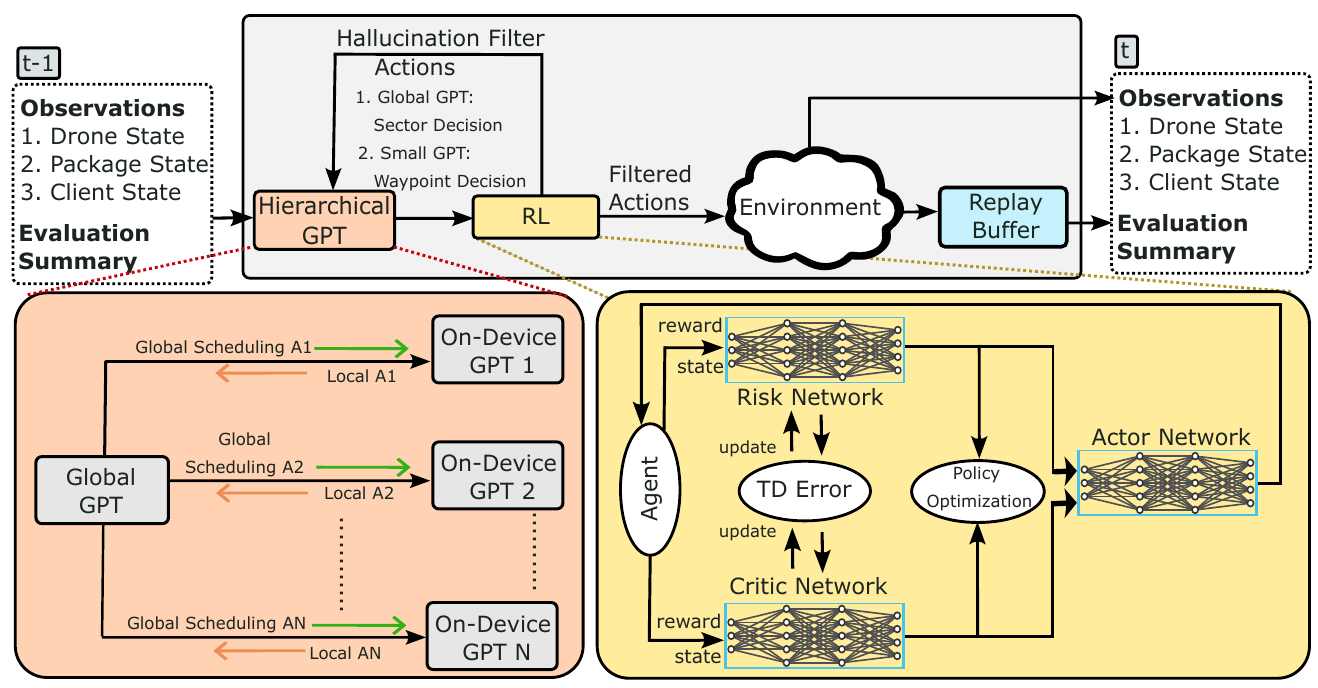}
    \caption{Two-tiered generative pretrained transformer architecture.}
    \label{fig:System_Arch}
    \vspace{-5mm}
\end{figure*}

\subsection{Generative Pretrained Transformer}
Within the drone-based logistics framework, a GPT is utilized for autonomous decision-making, operating through two distinct yet complementary modules: the Global GPT model and the on-device GPT model.

\BfPara{Global GPT Model} This model functions as a central decision-maker by processing detailed simulation state data—including drone statuses, sector-specific customer queues, and memory logs—to generate actions that direct drones to their designated areas. This process is formulated as a conditional probability distribution over actions. i.e., 
\begin{equation}
P(y_t \mid y_{<t}, X) = \mathrm{softmax}\Big(f_\theta(y_{<t}, X)\Big),
\end{equation}
where \( y_t \) denotes the action at time \( t \), \( y_{<t} \) represents the preceding tokens, and \( X \) is the comprehensive input context. The model employs an attention mechanism is defined as,
\begin{equation}
\mathrm{Attention}(Q, K, V) = \mathrm{softmax}\left(\frac{QK^T}{\sqrt{d_k}}\right)V,
\end{equation}
where this is for extracting salient features from the input, thereby enabling robust global planning and decision-making.

\BfPara{On-Device GPT Model} In contrast, the on-device GPT model is designed for deployment on individual drones or edge devices. It handles real-time, local decision-making by adapting the global strategy to immediate environmental conditions. Its lightweight structure ensures minimal latency and efficient computation, which is critical for in-situ adjustments. Additionally, it incorporates a dedicated mechanism for optimizing delivery routes by computing an efficient sequence that minimizes travel costs (e.g., distance or battery consumption) between consecutive drop-off points expressed as,
\begin{equation}
\min_{\pi} \sum_{i=1}^{N}\nolimits c\big(\pi(i), \pi(i+1)\big),
\end{equation}
where \( \pi \) represents the permutation of delivery sequences and \( c\big(\pi(i), \pi(i+1)\big) \) denotes the associated travel cost.

This approach, combining global strategic oversight with local adaptive control, enhances both the efficiency and robustness of autonomous logistics systems, as shown in Fig.~\ref{fig:System_Arch}.


\subsection{Reinforcement Learning with Safety Guarantees}\label{subsection:SRL4contact}
An RL framework that integrates inherent safety guarantees for the drone-based logistics system is introduced. This framework is formulated as a constrained Markov decision process (CMDP), which is defined as,
   $CMDP = (\mathcal{S}, \mathcal{A}, r, P, \gamma, \mu, \Omega)$,
where \(\mathcal{S}\) represents the state space, \(\mathcal{A}\) denotes the action space, \(r: \mathcal{S}\times\mathcal{A}\rightarrow\mathbb{R}\) is the reward function, \(P\) corresponds to the state transition probability, \(\gamma\in [0,1]\) is the discount factor, \(\mu\) signifies the initial state, and \(\Omega\) are the set of constraints that must be satisfied.
In the context of drone logistics, all vectors are defined in a global reference frame. Specifically:
\begin{itemize}
    \item \textit{State \(\mathcal{S}\):} comprises the normalized drone position, battery level, sector-specific customer counts, current sector assignments, and accumulated travel distance.
    \item \textit{Action \(\mathcal{A}\):} consists of commands that assign drones to operational sectors and toggle between idles and transits.
    \item \textit{Reward \(r\):} is designed to balance delivery performance with penalties for excessive travel, high battery consumption, and violations of safety criteria.
    \item \textit{Constraints \(\Omega\):} enforce operational requirements such as battery preservation, balanced sector allocation, avoidance of redundant customer visits, and route efficiency, with each constraint modeled as a cost function \(c(s,a)\).
\end{itemize}

Safety is embedded into the RL framework by employing a Lagrangian dual formulation that aligns reward maximization with adherence to constraints is expressed as,
\begin{multline}
\mathcal{L}(\pi, \lambda) = \mathbb{E}_{\pi}\left[\sum_{t=0}^{\infty}\nolimits \gamma^t r(s_t,a_t)\right] \\ - \lambda^\top\left(\mathbb{E}_{\pi}\left[\sum_{t=0}^{\infty}\nolimits \gamma^t c(s_t,a_t)\right] - b\right), \label{eq:lagrangian}
\end{multline}
where \(\lambda\ge 0\) serves as the Lagrange multiplier and \(b\) specifies the designated safety thresholds.

Moreover, a safety value function is defined to quantify the cumulative cost under policy \(\pi\) expressed as,
\begin{equation}
V_{c}^{\pi}(s) = \mathbb{E}_{\pi}\left[\sum_{t=0}^{\infty}\nolimits\gamma^t c(s_t,a_t) \,\bigg|\, s_0=s\right]. \label{eq:safevalue}
\end{equation}
 
Within the actor-critic architecture, the actor network generates actions while the critic network evaluates both the expected return \(Q^{\pi}(s,a)\) and the associated safety cost \(Q_{c}^{\pi}(s,a)\). Consequently, the policy gradient is updated as follows,
\begin{multline}
\nabla_{\theta}J(\theta) = \mathbb{E}_{s\sim d^{\pi},\,a\sim\pi_{\theta}}\Big[\nabla_{\theta}\log\pi_{\theta}(a|s)\\ \times \left(Q^{\pi}(s,a)-\lambda\,Q_{c}^{\pi}(s,a)\right)\Big], \label{eq:policygrad}
\end{multline}
where \(\theta\) is the policy parameters.
To maintain operational safety, an override mechanism is introduced. The selected action incur a cost that exceeds the safety threshold (\(c(s_t,a_t)>b\)), a safe alternative \(\hat{a}_t\) replaces the original action is as,
\begin{equation}
a_t^{\text{safe}} = 
\begin{cases}
a_t, & \text{if } c(s_t,a_t)\le b, \\
\hat{a}_t, & \text{otherwise.}
\end{cases} \label{eq:override}
\end{equation}
Fig.~\ref{fig:System_Arch} illustrates the RL network architecture with integrated safety measures, detailing the interactions among the safety critic, recovery policy modules, and the primary task policy, which undergoes continual refinement.

\vspace{-2mm}
\section{Algorithm Design}
\label{sec:algorithm_design}

\subsection{GPT-Driven Planning and State Abstraction}
A two-tier GPT-based strategy is adopted, i.e., \textbf{Global GPT (High-Level Planner)} (This component allocates UAVs to operational sectors or primary tasks (e.g., \textit{"Drone~1, proceed to Sector~East"})) and \textbf{On-Device GPT (Local Route Planner)} (Upon assignment at the high level, this formulates detailed routes for visiting specified customer locations (e.g., \textit{"Visit Customers A, B, and C sequentially, then return to base"})).

In the implementation, the Global GPT is instantiated using the GPT-4o model, while the On-Device GPT is deployed with the 4o-mini model. Let \(N\) denote the number of UAVs. The system state at \(t\) is defined as,
$s_t = \Bigl(\mathbf{x}(t),\, \mathbf{b}(t),\, \mathbf{q}(t)\Bigr)$,
where \(\mathbf{x}(t)=\{x_i(t)\}_{i=1}^N\) indicates the positions or sector assignments of UAVs, \(\mathbf{b}(t)=\{b_i(t)\}_{i=1}^N\) represents their battery levels, and \(\mathbf{q}(t)\) comprises the set of pending delivery demands.

The Global GPT module generates an action for each UAV \(i\) at time \(t\) is expressed as,
$\tilde{a}_t^i = \mathrm{GPT}\Bigl(s_t,\, \mathrm{Memory}_t\Bigr)$,
where \(\mathrm{Memory}_t\) is a context buffer that holds recent decisions and state transitions. The collective proposed action vector is denoted by \(\tilde{\mathbf{a}}_t = \bigl(\tilde{a}_t^1,\dots,\tilde{a}_t^N\bigr)\).

\subsection{RL Constraints, Overriding, and Objective Function}
To ensure safe operation, a set of feasibility constraints \(\Omega = \{\Omega_1,\dots,\Omega_K\}\) is enforced, with each constraint defined as,
$g_k\bigl(s_t, \tilde{a}_t^i\bigr) \le 0,\quad \forall\, k \in \{1,\dots,K\}$,
which encapsulates requirements such as battery limits and no-fly zones. In instances where any constraint is violated (i.e., if \(g_k\bigl(s_t, \tilde{a}_t^i\bigr)>0\) for any \(k\)), the RL module substitutes the proposed action with a safe fallback \(\hat{a}_t^i\), such that,
\begin{equation}
a_t^i =
\begin{cases}
\tilde{a}_t^i, & \text{if } g_k\le 0\quad \forall\, k,\\[4pt]
\hat{a}_t^i, & \text{otherwise.}
\end{cases}
\label{eq:alg_design_action_override}
\end{equation}

The overall objective of the RL agent is to maximize the cumulative discounted reward while penalizing constraint violations and deviations from the GPT-proposed actions. Accordingly, the following objective function is defined as,
\begin{multline}
\max_{\pi} \quad \mathbb{E}\!\Biggl[
\sum_{t=0}^T\nolimits \gamma^t \Bigl( r\bigl(s_t, \mathbf{a}_t\bigr)\\[2mm]
 - \sum_{k=1}^K\nolimits \lambda_k \, \max\{0,\, g_k(s_t,\mathbf{a}_t)\}
- \eta\, \bigl\|\mathbf{a}_t - \tilde{\mathbf{a}}_t\bigr\|^2 \Bigr)
\Biggr],
\label{eq:alg_design_objective_modified}
\end{multline}
where \(\pi\) represents the policy that determines the final action \(a_t^i\), \(\lambda_k \ge 0\) are Lagrange multipliers that penalize constraint violations, and \(\eta > 0\) is a regularization parameter that discourages significant deviations from the GPT-suggested actions.

\begin{figure*}[t]
  \centering
  \small
  \begin{tcolorbox}[title=Planning Prompt for Global GPT]
        You are in charge of a delivery drone in a logistics environment. Your task is to determine the optimal high-level action for the drone based on the current simulation state. Agents interact in environments via programs which are instructions that describe which actions each agent should perform, and with which objects. Each decision should follow the format -\\ 
    Decision: \{action\_choice\} \\
    Some examples of decision-action pairs are given below - \\
    \{incontext\_examples\_global\}\\
    You have the following state information: \{state\_info\}. The list of available actions are - ["go\_to\_sector\_east", "go\_to\_sector\_west", "go\_to\_sector\_north", "go\_to\_sector\_south", "idle"]. Ensure that the selected action complies with operational safety constraints (e.g., battery limits, no-fly zones). If any constraint is likely to be violated, return <pass> instead of an unsafe decision.
    
    \textbf{Description}: \{input\} \\
    \textbf{Decision:}
  \end{tcolorbox}
  \caption{This prompt directs the Global GPT to generate high-level, safety-compliant decisions for drone sector allocation.}
  \label{fig:global-gpt-prompt}
\end{figure*}

\begin{figure*}[t]
  \centering
  \small
  \begin{tcolorbox}[title=Execution Prompt for On-Device GPT]
        You are in control of a delivery drone in a logistics simulation. Your task is to generate a single line of program in the specified format based on the provided route plan. Agents interact in environments via programs which are instructions that describe which actions each agent should perform, and with which objects. Each line of program must follow the format -\\ 
    Action Plan: \{command\_syntax\} \\
    Some examples of plan-action program pairs are given below - \\
    \{incontext\_examples\_execution\}\\
    You have the following objects in the scene: Warehouse located at \{warehouse\_location\} and Customers: \{customers\_list\}. The list of available actions are - ["move\_to\_customer", "return\_to\_base", "idle"]. Ensure that the generated command adheres to safety requirements; if any safety constraint may be violated, output <pass> rather than an unsafe action.
    
    \textbf{Description}: \{input\} \\
    \textbf{Action Plan}:
  \end{tcolorbox}
  \caption{This prompt directs the On-Device GPT to produce control commands that are consistent with the prescribed safety constraints for the drone route plan.}
  \label{fig:on-device-gpt-prompt}
    \vspace{-3mm}
\end{figure*}

\subsection{Dual Replay Mechanism and Integrated Algorithm}
To facilitate effective learning and decision refinement, we employ two replay buffers, i.e., \textbf{RL Replay Buffer} $\mathcal{B}_\mathrm{RL}$ (Stores tuples $(s_t,\mathbf{a}_t,r_t,s_{t+1})$, which are used to update the RL policy via standard actor-critic methods) and \textbf{GPT Replay Buffer} $\mathcal{B}_\mathrm{GPT}$ (Records entries $(s_t,\tilde{a}_t^i,\mathrm{override\_flag})$ to capture instances where GPT-proposed actions were overridden, thereby guiding the GPT to avoid unsafe proposals in future iterations).

In the integrated framework, each UAV first queries the Global GPT for a high-level action. The RL module then evaluates the proposed action against the safety constraints and, if necessary, substitutes it with a safe alternative as per \eqref{eq:alg_design_action_override}. The environment subsequently transitions to the next state, and both replay buffers are updated with the observed data. Periodic updates of the RL policy are performed using mini-batches sampled from $\mathcal{B}_\mathrm{RL}$, while the GPT module leverages $\mathcal{B}_\mathrm{GPT}$ to refine its future decision-making.
This integrated design enables scalable two-tier planning, where the Global GPT orchestrates high-level assignments and the on-device GPT refines local routes in real time. The RL layer provides formal safety guarantees by verifying constraints and overriding risky actions, thereby ensuring reliable and cost-effective UAV operations. Moreover, the dual replay mechanism fosters synergy between the semantic reasoning of the GPT and the constraint-driven RL, ultimately enhancing overall system performance.


\begin{figure*}
    \centering
    \includegraphics[width=0.4\linewidth]{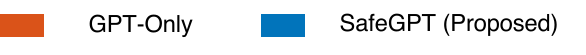}\\
    \vspace{-4mm}    
    \subfigure[Reward.]{
    \includegraphics[width=0.31 \linewidth]{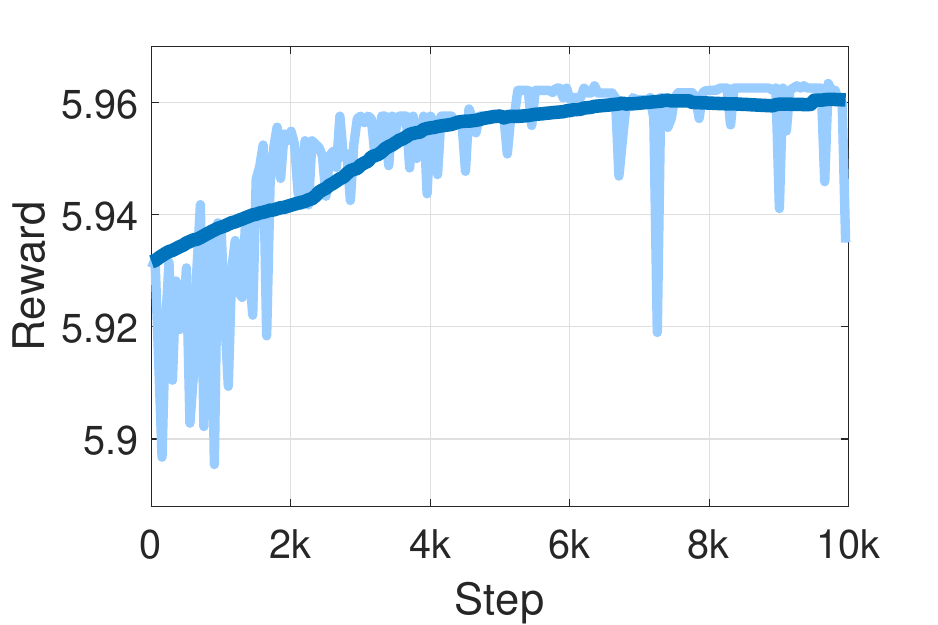}
    \label{fig: Reward}
    }
    \subfigure[Battery consumption.]{
    \includegraphics[width=0.31\linewidth]{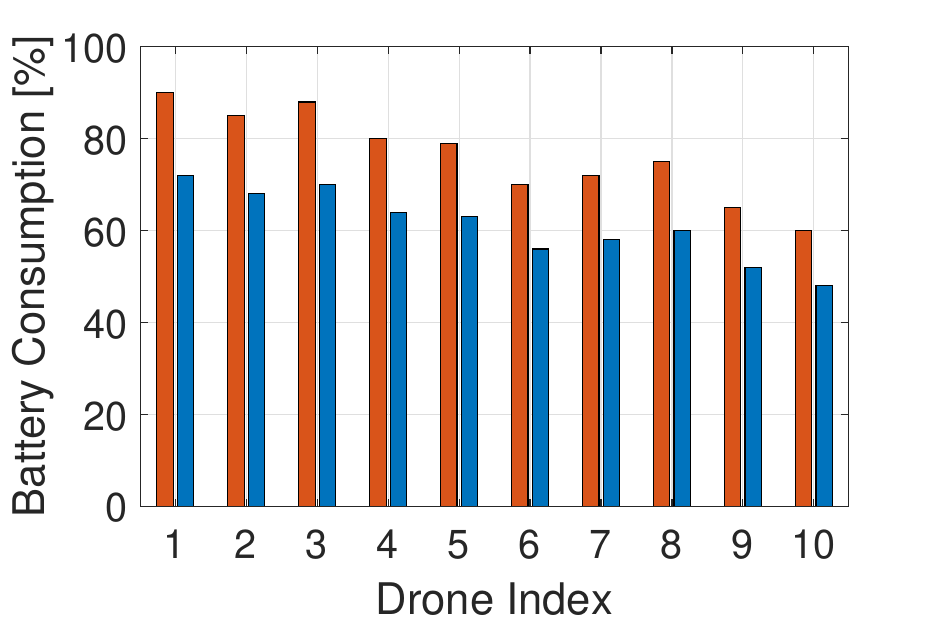}
    \label{fig:Battery Consumption}
    }
    \subfigure[Travel distance.]{
    \includegraphics[width=0.31 \linewidth]{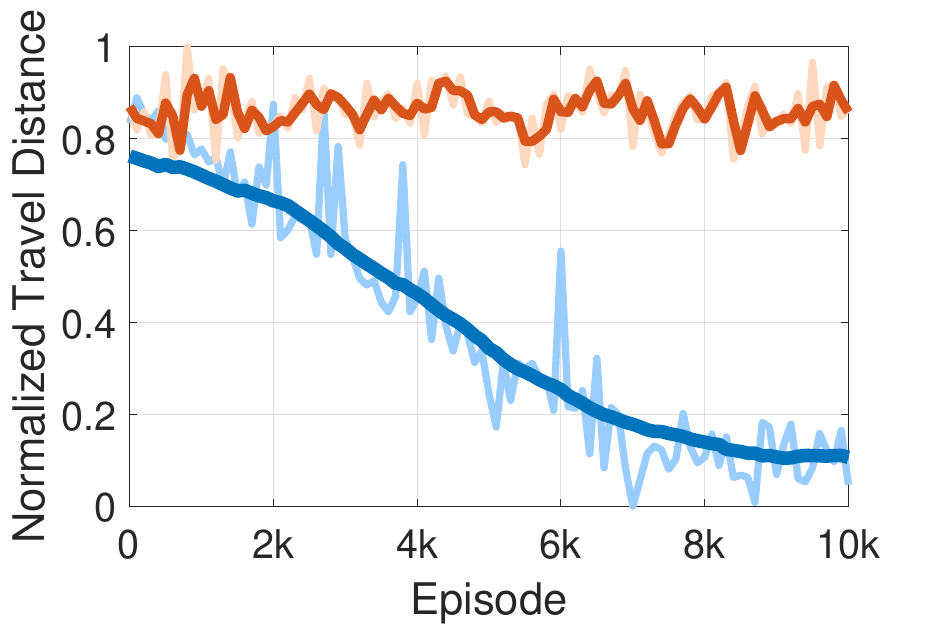}
    \label{fig:Travel Distance}
    }

    \caption{
Control performances of the UAV trained using the proposed algorithm and benchmark methods in dynamic environments.
}
    \label{fig:performance_control}
    \vspace{-3mm}
\end{figure*}


\section{Performance Evaluation}
\label{sec:V}

\subsection{Experimental Setup}


All experiments are conducted within a two-dimensional urban grid of \(1000\times1000\)\,meters, defined over the Cartesian range \([-500,500]\times[-500,500]\). A single warehouse, serving as the central depot, is positioned at the origin \((0,0)\). A total of 10 drones initiates each episode at the warehouse with a full battery. Drones become inactive once their battery is exhausted.

\paragraph{Episode Length and Customer Generation} 
Each simulation run is segmented into episodes comprising up to 300 time steps, where each time step represents a discrete decision interval (e.g., a few seconds). At the start of each episode, customers are stochastically generated within four primary sectors (\texttt{EAST}, \texttt{NORTH}, \texttt{WEST}, and \texttt{SOUTH}). 

\begin{table}[t!]
\footnotesize
\renewcommand{\arraystretch}{1.15}
\centering
\caption{Delivery success rate comparison.}
\begin{tabular}{lcc}
\toprule
\textbf{Method} & \textbf{Success Rate [\%]} \\
\midrule
GPT-only & 98 \\
SafeGPT  & 100 \\
\bottomrule
\end{tabular}
\label{tab:delivery_count}
    \vspace{-2mm}
\end{table}

\begin{table}[t!]
\centering
\footnotesize
\setlength{\tabcolsep}{4pt}
\renewcommand{\arraystretch}{0.9}
\caption{Battery consumption for each drone.}
\begin{tabular}{lcccccccccc}
\toprule
Method / Agent & 1  & 2  & 3  & 4  & 5  & 6  & 7  & 8  & 9  & 10 \\
\midrule
GPT-only    & 90 & 85 & 88 & 80 & 79 & 70 & 72 & 75 & 65 & 60 \\
SafeGPT     & 72 & 68 & 70 & 64 & 63 & 56 & 58 & 60 & 52 & 48 \\
\bottomrule
\end{tabular}
\label{tab:raw_data_horizontal}
\vspace{-2mm}
\end{table}

\begin{table}[t!]
\centering
\footnotesize
\setlength{\tabcolsep}{4pt}
\renewcommand{\arraystretch}{0.9}
\caption{Summary statistics for battery consumption.}
\begin{tabular}{lcc}
\toprule
Method   & Mean battery consumption & STD battery consumption \\
\midrule
GPT-only & 76.4                     & 9.4                     \\
SafeGPT  & 61.1                     & 7.5                     \\
\bottomrule
\end{tabular}
\label{tab:summary_stats}
    \vspace{-3mm}
\end{table}

\paragraph{Comparison: GPT-Only vs.\ SafeGPT}
Two variants of the control strategy are evaluated, i.e., \BfPara{GPT-Only} (Drones rely solely on outputs from the GPT for both sector assignment and routing decisions. Although the GPT utilizes a replay buffer to maintain state-action context, no explicit RL-based safety module is implemented) and \BfPara{SafeGPT (Proposed)} (In this approach, the GPT continues to provide high-level decisions (e.g., sector targets and delivery routes). However, an RL module validates or overrides any action that breaches constraints).
This comparison elucidates the impact of real-time constraint on successful deliveries and resource utilization.

\paragraph{Simulation Procedure}
At the commencement of each episode, all drones are reset to the warehouse with a full battery. Customers are generated randomly at the beginning of each episode. During each time step, each active drone queries the decision-making module---either the standalone GPT or the combined SafeGPT---for an action. Drones may elect to idle, navigate toward a designated sector, or follow a precomputed route to execute deliveries. The episode terminates either when all customers have been served or upon reaching 300 time steps, after which the simulation resets for the subsequent episode. 

\paragraph{Performance Metrics}
The key outcome metrics evaluated include following factors, i.e., \textbf{Battery Consumption} (The average energy expended by drones prior to deactivation or the termination of an episode), \textbf{Travel Distance} (The cumulative distance traveled by each drone, serving as an indicator of route efficiency), \textbf{Delivery Success Rate} (The total number of customers served per episode), and \textbf{Hallucination Count} (The frequency of unrealistic or unsafe actions suggested by the GPT, reported explicitly for the SafeGPT approach and noted as dangerous actions in the GPT-only setup).
As illustrated in Fig.~\ref{fig:global-gpt-prompt} and Fig.~\ref{fig:on-device-gpt-prompt}, the framework employs carefully constructed prompts for the Global GPT (high-level planning) and the On-Device GPT (local execution). These prompts guide the GPT to generate valid, semantically rich plans while minimizing spurious or irrelevant outputs. Fig.~\ref{fig:performance_control} compares the control performance of both GPT-only and SafeGPT.

\begin{table}[t!]
\footnotesize
\centering
\caption{Safety constraints and counts.}
\begin{tabularx}{0.5\linewidth}{l c}
\toprule[1pt]
\textbf{Safety constraint} & \textbf{Count [\%]} \\
\midrule
Duplicate visits &
$0.64$\; \tikz{
\draw[gray,line width=.3pt] (0,0) -- (1.1,0);
\draw[white,line width=0.01pt] (0,-2pt) -- (0,2pt);
\draw[black,line width=1pt] (0.640,0) -- (0.740,0);
\draw[black,line width=1pt] (0.640,-2pt) -- (0.640,2pt);
\draw[black,line width=1pt] (0.740,-2pt) -- (0.740,2pt);
} \\
Battery safety &
$0.24$\; \tikz{
\draw[gray,line width=.3pt] (0,0) -- (1.1,0);
\draw[white,line width=0.01pt] (0,-2pt) -- (0,2pt);
\draw[black,line width=1pt] (0.240,0) -- (0.340,0);
\draw[black,line width=1pt] (0.240,-2pt) -- (0.240,2pt);
\draw[black,line width=1pt] (0.340,-2pt) -- (0.340,2pt);
} \\
Route efficiency &
$0.08$\; \tikz{
\draw[gray,line width=.3pt] (0,0) -- (1.1,0);
\draw[white,line width=0.01pt] (0,-2pt) -- (0,2pt);
\draw[black,line width=1pt] (0.080,0) -- (0.180,0);
\draw[black,line width=1pt] (0.080,-2pt) -- (0.080,2pt);
\draw[black,line width=1pt] (0.180,-2pt) -- (0.180,2pt);
} \\
Sector balance &
$0.04$\; \tikz{
\draw[gray,line width=.3pt] (0,0) -- (1.1,0);
\draw[white,line width=0.01pt] (0,-2pt) -- (0,2pt);
\draw[black,line width=1pt] (0.040,0) -- (0.140,0);
\draw[black,line width=1pt] (0.040,-2pt) -- (0.040,2pt);
\draw[black,line width=1pt] (0.140,-2pt) -- (0.140,2pt);
} \\
\bottomrule[1pt]
\end{tabularx}
\label{tab:safety_constraints}
    \vspace{-5mm}
\end{table}

\subsection{Results of Experiment}
\label{sec:results_exp}
The effectiveness of the proposed SafeGPT approach is examined in comparison to a baseline method that relies entirely on GPT-based control. 
Table~\ref{tab:delivery_count} presents the overall success rates for both GPT-only and SafeGPT, where the success rate is defined as the fraction of customers who receive their deliveries within the maximum episode duration.
SafeGPT achieves a 100\% delivery rate, surpassing GPT-only with 98\%. The RL component prevents any missed deliveries that could occur under dynamic constraints such as limited battery capacity or strict deadlines. Fig.~\ref{fig:performance_control}(a) illustrates the evolution of the training reward for SafeGPT as learning proceeds. Note that training is applied exclusively to SafeGPT, as GPT-only does not incorporate reinforcement learning; SafeGPT begins to converge around 7000 episodes.
Fig.~\ref{fig:performance_control}(b) depicts the average battery depletion per drone after completing all missions. An analysis using Table~\ref{tab:raw_data_horizontal} and Table~\ref{tab:summary_stats} reveals that under the GPT-only method, battery consumption per drone ranges from 60{\%} to 90{\%}, whereas SafeGPT achieves lower consumption, with values ranging from 48{\%} to 72{\%}. Moreover, the mean battery consumption for GPT-only is 76.4{\%} (std: 9.4), compared to SafeGPT's mean of 61.1{\%} (std: 7.5), indicating efficient and uniform energy usage.
Fig.~\ref{fig:performance_control}(c) compares the total travel distance over 10,000 training episodes, illustrating that SafeGPT reduces route length more effectively than GPT-only as training progresses. By episode 10,000, SafeGPT exhibits significantly shorter paths due to the override mechanism that mitigates unnecessary detours and steers drones directly toward their targets.
Regarding hallucinations, the frequency of infeasible or unsafe actions suggested by the GPT is recorded. As shown in Table~\ref{tab:safety_constraints}, the most frequent issues arise from duplicate visits, followed by battery safety violations, inefficient routing, and sector imbalance. In the SafeGPT approach, such problematic instructions are automatically overridden by the RL module, thereby substantially reducing hallucination-induced deviations.

\section{Concluding Remarks} \label{sec:conclusion}
SafeGPT provides a concise two-tiered planning strategy that unites GPT reasoning with RL and dual replay buffers for UAV-based logistics. The framework assigns tasks globally and locally via GPT modules, while RL overrides unsafe actions that violate constraints such as battery preservation and route efficiency, using dual replay buffers to refine decisions from past experiences. Experimental evaluations in UAV delivery environment demonstrate reductions in travel distance and battery usage, as well as the prevention of repeated customer visits and other infeasible routes, offering a promising approach for reliable and cost-effective unmanned delivery services.

\bibliographystyle{IEEEtran}
\bibliography{ref_quantum}
\end{document}